\documentclass[letterpaper,10pt,conference]{IEEEtran}
\IEEEoverridecommandlockouts
\usepackage{cite}
\usepackage{amsmath,amssymb,amsfonts}
\usepackage{algorithmic}
\usepackage{graphicx}
\usepackage{textcomp}
\usepackage{xcolor}
\usepackage{hyperref}
\usepackage{booktabs}
\def\BibTeX{{\rm B\kern-.05em{\sc i\kern-.025em b}\kern-.08em
    T\kern-.1667em\lower.7ex\hbox{E}\kern-.125emX}}

\begin{document}

\title{\vspace{1cm}SPARC: RoboCup Standard Platform for Autonomous Rescue Competition with Initial Datasets, Software, and Simulation Environments
}


\author{Yuezhong Wang$^{1,*}$, Bichi Zhang$^{1,*}$, Zhongli Yin$^{1}$, Pingcong Li$^{1}$, Raimund Edlinger$^{2}$ and S\"oren Schwertfeger$^{1,\dagger}$%
\thanks{$^{1}$ The Key Laboratory of Intelligent Perception and Human-Machine Collaboration - ShanghaiTech University, Ministry of Education, China.}%
\thanks{$^{2}$ Smart Automation and Robotics, University of Applied Sciences Upper Austria, Wels, Austria.}%
\thanks{$^{*}$Equal contribution.}%
\thanks{$^{\dagger}$Corresponding author.}%
\thanks{This work was supported by the National Natural Science Foundation of China under Grant W2531052 and the Key Laboratory of Intelligent Perception and Human-Machine Collaboration (ShanghaiTech University), Ministry of Education, China. The experiments were supported by the Core Facility Platform of Computer Science and Communication, SIST, ShanghaiTech University.}%
}

\maketitle

\begin{abstract}
The RoboCup Standard Platform for Autonomous Rescue Competition (SPARC) is a new competition under development of the RoboCup Rescue Robot League that uses the Unitree A2-W as a common robot platform. SPARC aims to make research on artificial intelligence for autonomous rescue robots accessible to a broader research community, including groups without expertise in developing and maintaining custom robot hardware. By providing a capable standard platform and established benchmark environments, researchers can focus on autonomous navigation, perception, manipulation, and embodied AI for challenging disaster-response scenarios. This paper introduces the SPARC competition, its benchmark environment, and the Unitree A2-W standard platform. Furthermore, we present the first open SPARC datasets collected during the IEEE ICRA 2026 Legged Robot Challenges and RoboCup Rescue 2026, release an open-source software stack for autonomous navigation, and provide initial simulation assets representing RoboCup Rescue test environments. Together, these resources aim to lower the barrier to entry for autonomous rescue robotics and facilitate reproducible research and collaborative development on a common hardware platform.
\end{abstract}


\section{Introduction}

Natural and man-made disasters continue to pose significant challenges for emergency responders. Search and rescue robots are increasingly being deployed in these scenarios \cite{surmann2024lessons}. Consequently, rescue robotics has become an important research area \cite{schneider2016assessing}. Owing to the wide range of capabilities required, including robust mobility, autonomous navigation, perception, mapping, manipulation, and human-robot interaction, disaster response has been recognized as one of the most realistic and comprehensive benchmarks for autonomous robotic systems \cite{hu2026shifting}.

For more than two decades, the RoboCup Rescue Robot League has served as a benchmark competition for autonomous rescue robotics \cite{sheh2011robocuprescue,sheh2013advancing,sheh201616}. By providing realistic disaster scenarios together with objective evaluation methodologies, the league has driven research in autonomous navigation, mapping, perception, manipulation, and human-robot interaction. However, participating teams have traditionally developed and maintained their own robotic platforms, resulting in duplicated engineering effort and making reproducible comparisons between algorithms challenging. So, while RoboCup Rescue has successfully standardized benchmark environments and evaluation procedures, the participating robots have remained highly heterogeneous.

\begin{figure}[tb]
    \centering
    \includegraphics[width=0.99\linewidth]{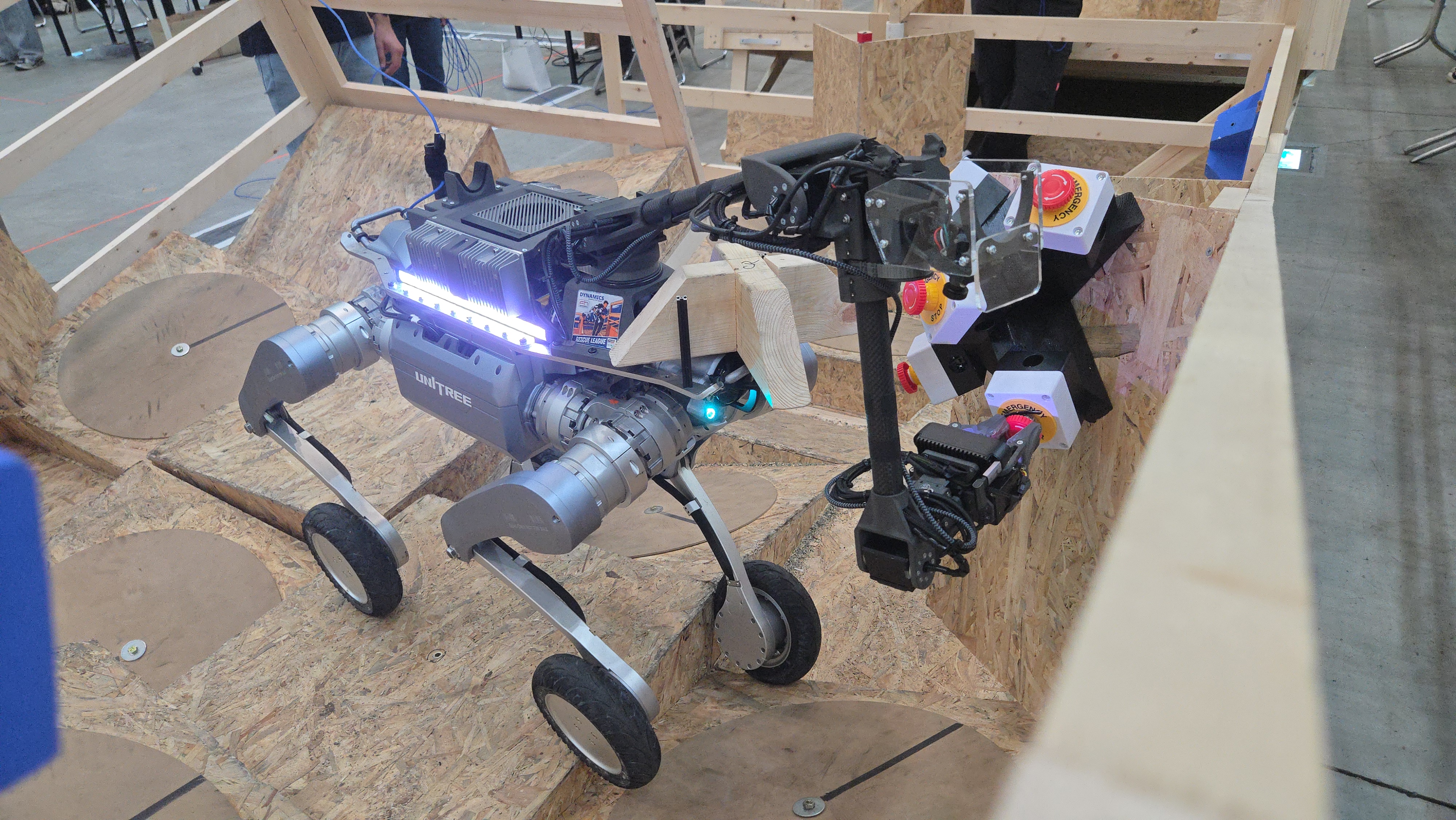}
    \caption{Unitree A2-W Pro configured for teleoperated and autonomous rescue operation during RoboCup Rescue 2026, showing the standard mobility platform together with the research-specific sensing, computing, and manipulation components.}
    \label{fig:A2-W}
\end{figure}

A successful solution to this problem has already been demonstrated within RoboCup. In the RoboCup Soccer Standard Platform League, all teams compete using identical robot hardware, allowing research efforts to focus on perception, localization, decision making, multi-robot cooperation, and artificial intelligence rather than hardware development \cite{chown2014standard,genter2017three}. This approach has lowered the barrier to entry, improved reproducibility, and accelerated software development through shared tools, datasets, and open-source software.

\begin{figure}[t]
    \centering
    \includegraphics[width=\linewidth]{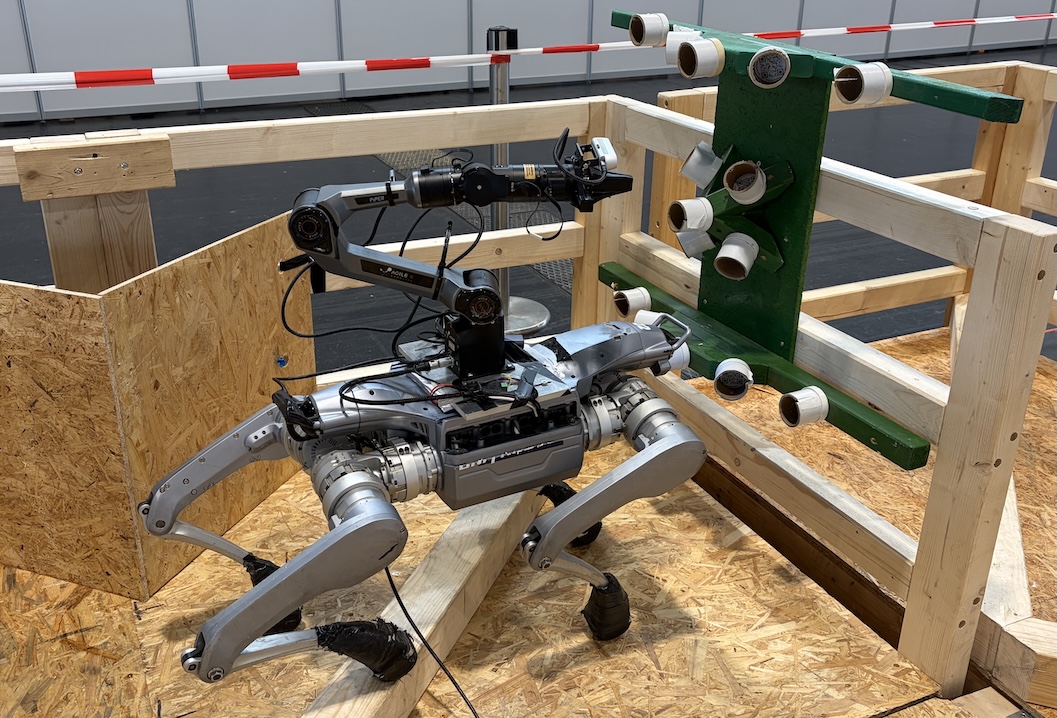}
    \caption{A Unitree A2 robot with an AgileX PiPER manipulator at the IEEE ICRA 2026 Legged Robot Challenges (LRC) in the \emph{K-Rails} lane performing an inspection task during the dexterity phase of the run.}
    \label{fig:A2}
\end{figure}

Recent advances in commercially available quadruped and wheel-legged robots have made a standard platform feasible for rescue robotics. Platforms such as the Unitree A2-W combine high mobility, integrated sensing, and onboard computation at a price point accessible to many research groups. Their capabilities have already been demonstrated in recent competitions, with an A2-W winning RoboCup Rescue 2026 (see Fig.~\ref{fig:A2-W}) and the closely related A2 platform winning the Best in Class Autonomy award at the IEEE ICRA Legged Robot Challenge (LRC) 2026 (see Fig.~\ref{fig:A2}). These developments motivated the RoboCup Rescue League to establish the \emph{Standard Platform for Autonomous Rescue Competition (SPARC)}, using the Unitree A2-W as its standard robot and enabling researchers to focus on autonomous navigation, perception, manipulation, and embodied AI rather than hardware development.  SPARC is under active development, and contributions from the research community are encouraged. The latest competition rules, technical specifications, and updates can be found at \url{https://rrl.robocup.org/sparc/}.

This paper introduces the SPARC competition together with the first SPARC open research resources for the community, provided by the winner of both the RoboCup Rescue 2026 competition and the IEEE ICRA 2026 Legged Robot Challenges (LRC), and by the winner of the Best in Class Autonomy award of the ICRA 2026 LRC. The main contributions of this paper are:

\begin{enumerate}
    \item We introduce the RoboCup Standard Platform for Autonomous Rescue Competition (SPARC), including its motivation, competition format, benchmark tasks, evaluation methodology, and the Unitree A2-W standard platform.

    \item  We release the first public SPARC datasets, including data collected with the Unitree A2 Pro during the IEEE ICRA 2026 Legged Robot Challenges (LRC) and with the Unitree A2-W during RoboCup Rescue 2026 in Incheon, Korea.

    \item We open-source a complete software stack for autonomous navigation on the SPARC platform, enabling researchers to build upon state-of-the-art competition-winning solutions.

    \item     We publish initial simulation assets representing RoboCup Rescue test environments to support simulation-based development and evaluation.
\end{enumerate}

\begin{figure*}[t]
    \centering
    \includegraphics[width=\linewidth]{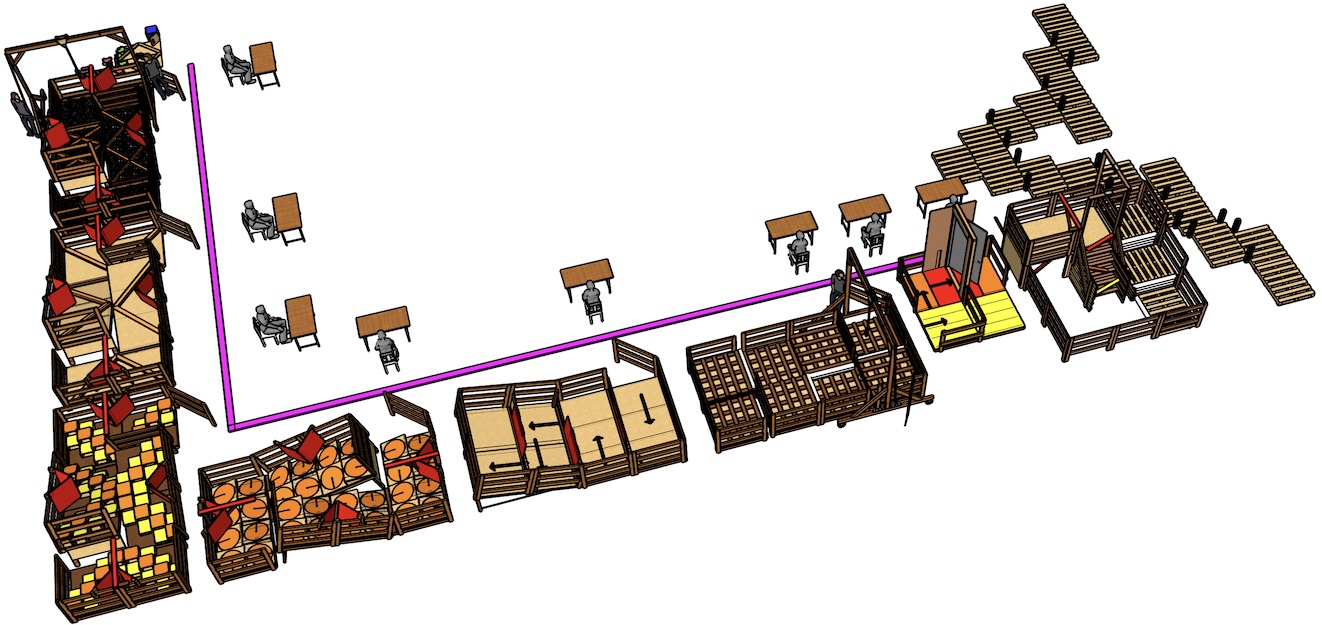}
    \caption{3D SketchUp model of the RoboCup Rescue 2026 arena in the more difficult, inclined configuration.}
    \label{fig:sketchup}
\end{figure*}

The remainder of this paper is organized as follows: Section~\ref{sec:related} reviews the related work. Section~\ref{sec:sparc} introduces the SPARC arena, the standard hardware platform, and the competition concept. The datasets released alongside this paper are presented in Section~\ref{sec:datasets}, while the open-source software stack is described in Section~\ref{sec:software}. Section~\ref{sec:simulation} introduces the simulation environment, and Section~\ref{sec:conclusions} concludes the paper.

\section{Related Work}
\label{sec:related}

\subsection{RoboCup Rescue}

The RoboCup Rescue Robot League was established to accelerate research in urban search and rescue by providing standardized benchmark environments that resemble real disaster scenarios \cite{sheh2011robocuprescue}. Over the years, the league has continuously evolved by introducing increasingly realistic environments, more challenging benchmark tasks, and improved evaluation methodologies \cite{sheh2013advancing,sheh201616}. Besides serving as an international competition, RoboCup Rescue has become an important experimental platform for evaluating autonomous navigation, mapping, perception, manipulation, and human-robot interaction under realistic operating conditions.

\subsection{Standard Platforms in RoboCup}

Standardized hardware platforms have proven highly successful in RoboCup Soccer. Rather than competing through custom robot hardware, all teams use identical robots and concentrate on software development, including perception, localization, decision making, and multi-robot coordination \cite{chown2014standard,genter2017three}. This approach has significantly lowered the entry barrier for new teams, improved reproducibility, simplified comparisons between algorithms, and encouraged open software development. SPARC adopts the same philosophy for rescue robotics, where recent advances in commercially available mobile platforms now make a standard platform practical.

\subsection{Rescue Robot Technologies}

Autonomous rescue robots require the integration of numerous capabilities. Robust mobility over challenging terrain remains one of the fundamental requirements, including autonomous stair climbing and traversal of complex obstacles \cite{chonnaparamutt2007fuzzy}. Reliable mapping and objective map evaluation have also received considerable attention, leading to evaluation methods based on topology graphs and structural similarity metrics \cite{schwertfeger2013evaluation}.

Environmental perception extends beyond navigation. Automatic recognition of hazardous material labels improves responder safety by allowing dangerous substances to be identified remotely \cite{edlinger2019hazmat}. Human-robot interaction remains equally important, particularly in communication-constrained environments where robust wireless networking and intuitive operator interfaces are essential \cite{edlinger2022intuitive}. Rescue robots are increasingly expected not only to locate victims but also to assist in casualty evacuation through autonomous manipulation and transportation systems \cite{edlinger2022innovative}.

Situational awareness further benefits from integrating multiple sources of information. Multi-layer map representations combine metric, semantic, and operational information into unified interfaces that support both robot autonomy and human decision making during rescue missions \cite{edlinger2024universal}.

\subsection{Research Gap}

Although considerable progress has been made in individual aspects of rescue robotics, existing research has largely focused on individual algorithms or custom robotic platforms. As a result, reproducible evaluation across different research groups remains difficult, and software reuse is limited. SPARC addresses this gap by introducing a common hardware platform, standardized benchmark environments, public datasets, simulation environments, and mandatory open-source software sharing. This enables reproducible comparisons while allowing researchers to concentrate on autonomous rescue algorithms rather than hardware development.

\section{RoboCup Standard Platform for Autonomous Rescue Competition (SPARC)}
\label{sec:sparc}

This section introduces the SPARC competition, including its benchmark test lanes, the standard robot platform, and the preliminary competition format and rules.

The RoboCup Standard Platform for Autonomous Rescue Competition (SPARC) is organized by the RoboCup Rescue League. To maximize compatibility with the established RoboCup Rescue competition, SPARC follows the open competition rules as closely as possible and adopts many of the same benchmark tasks and evaluation procedures. SPARC evaluates robotic systems using the same collection of standardized lanes as rescue, each designed to assess specific mobility, perception, and manipulation capabilities.

\subsection{RoboCup Rescue Lanes}

Fig.~\ref{fig:sketchup} shows a 3D model of the RoboCup Rescue 2026 competition arena as a representative example. From left to right, the arena consists of the lanes \emph{Gravel}, \emph{K-Rails}, \emph{Stepfield}, \emph{Rotating Ramps}, \emph{Center}, \emph{Pipes and Pallets}, \emph{Doors}, \emph{Stairs}, and \emph{Avoid}. A tenth lane, \emph{Maze}, is not shown.

The first six lanes share a common layout consisting of a $2.4\,\mathrm{m}\times4.8\,\mathrm{m}$ area formed by two L-shaped sections. To increase the difficulty, the central $1.2\,\mathrm{m}\times2.4\,\mathrm{m}$ sections of the first four lanes may be inclined. Optional pinch points (shown in red in Fig.~\ref{fig:sketchup}) further restrict the available space while guaranteeing a minimum clearance of $90\,\mathrm{cm}$.

Many of these lanes are derived from the standardized response robot test methods developed by ASTM Committee E54.09 on Response Robots \cite{astmE5409}, chaired by Adam Jacoff of the U.S. National Institute of Standards and Technology (NIST) \cite{messina2006performance}. By building upon these internationally recognized test methods, RoboCup Rescue provides benchmark scenarios that are both representative of real disaster-response tasks and reproducible across different competitions and research groups.

Each lane focuses on a particular set of robot capabilities, such as traversing uneven terrain, negotiating confined spaces, operating doors, climbing stairs, or performing manipulation tasks. Together, the lanes evaluate the complete robotic system, including locomotion, perception, localization, planning, manipulation, and system robustness, rather than isolated subsystem performance.

The \emph{Maze} lane consists of a modular arrangement of $1.2\,\mathrm{m}\times1.2\,\mathrm{m}$ cells with walls up to $2.4\,\mathrm{m}$ high. It is designed to evaluate autonomous exploration, simultaneous localization and mapping (SLAM), navigation, and object detection in confined environments.

The IEEE RAS Legged Robot Challenge (LRC), formerly known as the Quadruped Robot Challenge (QRC) \cite{jacoff2023taking}, adopts a simplified subset of the RoboCup Rescue benchmark tasks. Since its introduction at ICRA 2023, the challenge has been held at both ICRA and IROS and has become an important benchmark for autonomous legged robots. Compared with RoboCup Rescue, the LRC employs fewer lanes and simplified rules while preserving the same philosophy of standardized and reproducible evaluation. The close relationship between the two competitions has strongly influenced the design of SPARC.

\subsection{The SPARC Robot: Unitree A2-W}

The SPARC mobile base is the Unitree A2-W Pro (see Fig.~\ref{fig:A2-W}), a quadrupedal robot equipped with wheels on all four legs. It combines high mobility in difficult terrain with high-speed locomotion on easier surfaces. Its payload capacity, integrated sensor suite, onboard computation, and endurance make it well suited for rescue robotics. The RoboCup Rescue League selected the A2-W as the SPARC standard platform because it was, following the smaller Unitree Go2-W, one of the first commercially available wheel-legged robots while remaining sufficiently compact to operate within the existing RoboCup Rescue arena.

The robot features two onboard computers, one of which, equipped with an Intel Core i7 processor, is fully available for user applications. The 16-DoF robot integrates two Hesai JT128 3D LiDAR sensors with IMUs, a front-facing HD camera, bidirectional audio, and front lights. Including the batteries, the robot weighs approximately 52\,kg. With a payload of 25\,kg, it provides an operating time of approximately 1.5 hours and a range of up to 15\,km. 

The two LiDAR sensors are installed at the front and rear of the mobile base. Each 128-channel sensor provides an ultra-wide $360^\circ \times 189^\circ$ field of view and a maximum range of 60\,m. The LRC dataset presented in this paper was collected using the closely related Unitree A2 Pro, which provides the same sensor hardware and software interfaces but uses conventional feet instead of wheels.

Currently, the manipulator for the SPARC platform is not standardized; therefore, the initial competition will likely place greater emphasis on navigation tasks. The A2 robot used at the LRC was equipped with an AgileX PiPER manipulator, while the A2-W used at RoboCup Rescue 2026 carried a custom-built manipulator. 

\subsection{Competition Configuration of the Unitree A2-W}
The Unitree A2-W provides the standardized mobility platform for SPARC, while the integration of research-specific payloads remains open to participating teams. To evaluate the platform under realistic rescue conditions, a competition configuration of the A2-W was deployed during RoboCup Rescue 2026. The configuration retained the standard mobility base, onboard sensing, and low-level locomotion interface of the A2-W while extending the platform with additional hardware for autonomous navigation and manipulation.

The system was equipped with a lightweight articulated manipulator for interaction with the environment and manipulation-related rescue tasks. The manipulator was mounted directly on the robot body and operated independently of the low-level locomotion controller. This configuration allows navigation and manipulation capabilities to be evaluated on the same mobile platform while preserving the original mobility functionality of the A2-W.

The perception and navigation system uses the onboard LiDAR sensors and inertial measurements as the primary sources for state estimation and environment reconstruction. Additional computing resources are used for processing sensor data and executing the autonomous navigation stack. The resulting system architecture separates the platform-dependent components, such as locomotion and hardware interfaces, from the higher-level research components for state estimation, mapping, planning, and task execution.

\subsection{The SPARC Competition}

SPARC will operate in close collaboration with the traditional RoboCup Rescue competition. Following two days of setup and testing, RoboCup Rescue teams participate in two days of Preliminary rounds, typically completing a total of four runs in each of the ten lanes. The highest-ranking teams then advance to a one-day Semi-Final, in which selected lanes are combined into four longer sequences. A smaller number of teams subsequently compete in the Final, which follows a similar format for approximately half a day. At RoboCup Rescue 2026, 23 teams participated in the competition.

The exact SPARC rules are still under development by the RoboCup Rescue community, but several aspects of the competition format are already taking shape. During the Preliminary rounds, SPARC is expected to operate alongside the traditional RoboCup Rescue competition, using the same lanes and competition schedule. For the Semi-Finals, SPARC is expected to separate from the traditional competition and operate in an adjacent area, thereby also providing additional Semi-Final time slots for RoboCup Rescue. One possible exception is the \emph{Maze}, which SPARC may continue to share with RoboCup Rescue to evaluate autonomous exploration, SLAM, navigation, object detection, and search. This would require an appropriate time-sharing mechanism between the two competitions.

Current concepts for the SPARC Semi-Finals include collaborative tasks in which robots from two teams must jointly accomplish a common objective. Possible examples include coordinated pulling or transporting wooden blocks over longer distances and subsequently using them to construct temporary shoring structures, reflecting the use of timber shoring in real-world rescue operations \cite{usarShoringGuide}. Such tasks would extend the competition beyond the evaluation of individual robots toward multi-robot coordination, cooperation, and task planning in complex rescue scenarios.

The published SPARC scoring concept emphasizes autonomy, perception, and software capabilities, but retains teleoperation points at a lower relative weight. Accordingly, autonomous and teleoperated task execution must be reported separately rather than treating SPARC as an exclusively autonomous scoring system.

Establishing SPARC offers several advantages:
\begin{enumerate}
    \item By providing a standard robotic platform, SPARC enables research groups with limited expertise in robot hardware development to conduct research on autonomous rescue robotics. This lowers the barrier to entry and may attract researchers from artificial intelligence and related fields, potentially introducing new methods and approaches to rescue robotics.

    \item Since all SPARC teams use the same mobility platform, differences in competition performance can be more directly attributed to software and embodied AI capabilities. This enables more meaningful benchmarking, improves reproducibility, and facilitates repeatable experimental evaluation across different research groups.

    \item The use of common hardware and software interfaces substantially increases the value of sharing and reusing open-source software. SPARC will therefore require participating teams to release their software after each RoboCup World Championship. The resulting shared software ecosystem is expected to accelerate development and provide increasingly capable baseline systems for new and existing teams.

    \item Common mobility hardware and software interfaces also facilitate the development of standardized simulation environments. Combined with recent advances in robotics simulation, this allows research groups without access to the physical Unitree A2-W platform to develop, test, and benchmark their autonomous systems entirely in simulation and participate in virtual parts of the competition. At the physical competition, these teams may then deploy and evaluate their software on A2-W robots provided by the competition organizers. This model further lowers the barrier to participation by enabling research groups to contribute to autonomous rescue robotics without requiring them to purchase or maintain the physical robot hardware.
\end{enumerate}

\section{The 2026 SPARC Datasets} 
\label{sec:datasets}

Alongside this paper, we release sensor datasets collected with A2 and A2-W robots in real competition arenas, as well as two point clouds of one of the environments. The core sensor data are provided as ROS~2 bags recorded directly from the onboard sensors and robot-state interfaces. These datasets are intended to support the development, evaluation, and comparison of robotics algorithms, including localization, SLAM, mapping, and planning.

\begin{figure}[t]
    \centering
    \includegraphics[width=\linewidth]{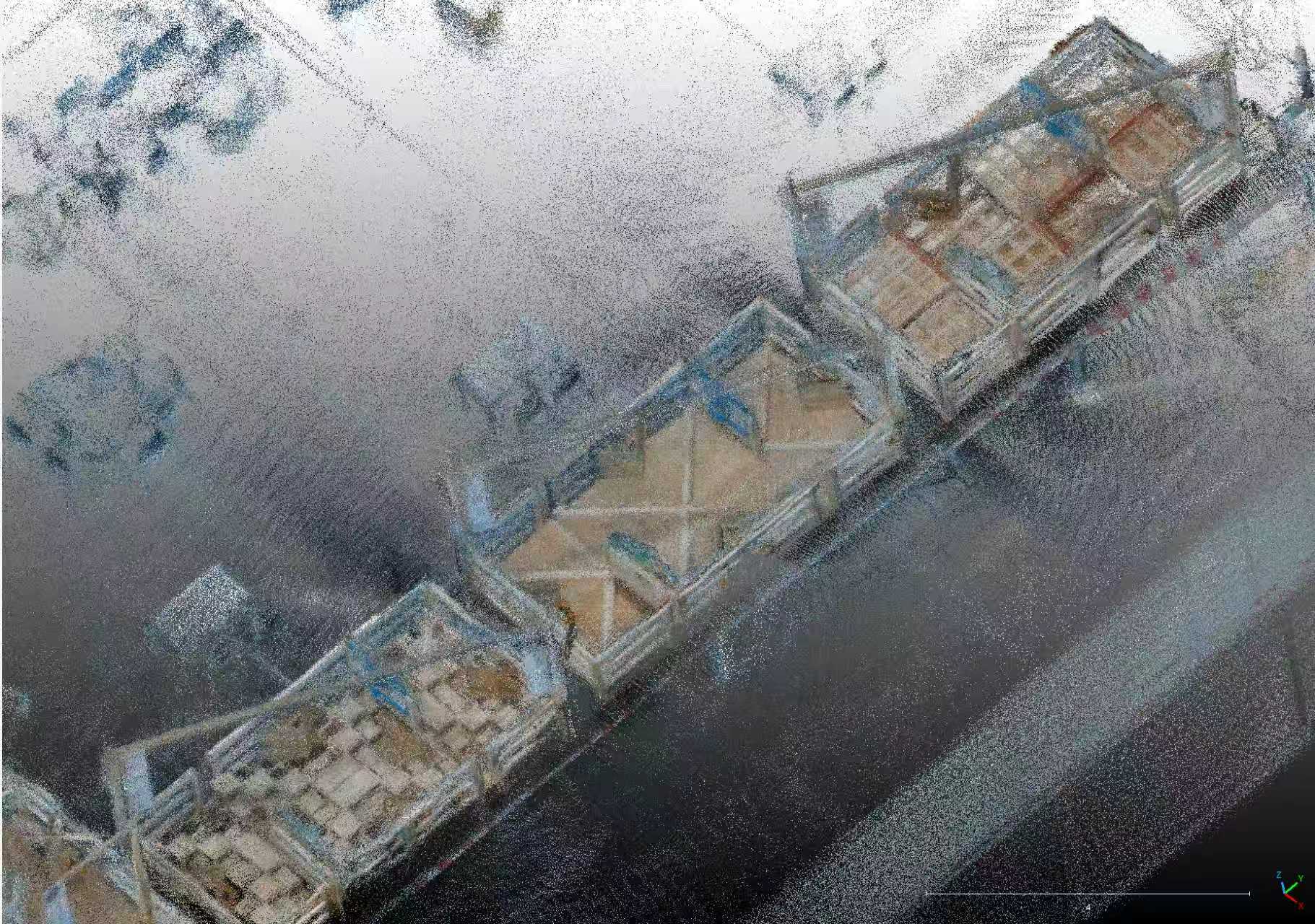}
    \caption{View of the IEEE ICRA 2026 LRC point cloud, showing from left to right the \emph{Stepfield}, the \emph{K-Rails} and the \emph{Pipes and Pallets} lanes.}
    \label{fig:LRC_pc}
\end{figure}

The datasets released with SPARC originate from two closely related competition platforms. The first dataset was collected using a Unitree A2 Pro during the IEEE ICRA 2026 Legged Robot Challenges (LRC) and contains two sequences from representative inclined benchmark lanes used throughout the competition. This robot doesn't feature the wheels on the legs, but the sensor suite and compute platform are the same. We also provide two colored point clouds of the LRC arena, collected with a hand-held 3D scanner. One was collected with the flat arrangement of the lanes (see Fig.~\ref{fig:LRC_pc}), while the second point cloud was scanned with the more difficult configuration used in the Semi-Finals, where the first four lanes were inclined. These high-resolution colored point clouds capture the geometry and appearance of the competition arena and are provided as map-level ground truth for evaluating maps reconstructed from the onboard sensor data. This dataset is open-sourced at \url{https://doi.org/10.5281/zenodo.23002951} \cite{wang_2026_23002951}. The second dataset was collected using the Unitree A2-W during RoboCup Rescue 2026 in Incheon, Korea, providing ROS~2 bag data from the official RoboCup Rescue competition environment and open-sourced at \url{https://dx.doi.org/10.21227/e01k-mf43}. 

The ROS~2 bags contain the onboard LiDAR, IMU, and robot-state measurements listed in Table~\ref{tab:datasets_LRC} and Table~\ref{tab:datasets_RC26}. The onboard LiDAR and IMU sensors are integrated into the Unitree A2/A2-W platform and accessed through the standard Unitree sensor interfaces. The recorded bags therefore preserve the platform-provided timestamps and coordinate-frame information used during data acquisition. No additional offline sensor calibration or external synchronization was performed during data collection; instead, the integrated sensor configuration and interfaces provided by the Unitree platform were used directly. The recorded topics are summarized in
Table~\ref{tab:datasets_LRC} and \ref{tab:datasets_RC26}, while the details of datasets are reported in Table~\ref{tab:data_details}.

\begin{table}[htbp]
\caption{SPARC Dataset Topics LRC 2026}
\begin{center}
\begin{tabular}{|l|c|}
\hline
\textbf{Topic}&\textbf{Description}\\
\hline
\texttt{/lidar\_points}  & Front LiDAR   \\
\texttt{/lidar\_imu}& Front LiDAR IMU   \\
\texttt{/lowstate}& Robot's Lowstate   \\
\texttt{/sportmodestate}& Robot's sportmodestate\\
\hline
\end{tabular}
\label{tab:datasets_LRC}
\end{center}
\end{table}

\begin{table}[htbp]
\caption{SPARC Dataset Topics RoboCup Rescue League 2026}
\begin{center}
\begin{tabular}{|l|c|}
\hline
\textbf{Topic}&\textbf{Description}\\
\hline
\texttt{/lidar\_points}  & Front LiDAR   \\
\texttt{/lidar\_points\_2}  & Rear LiDAR   \\
\texttt{/imu/data}& Microstrain IMU   \\
\texttt{/joint\_states}& Robot's joint configurations   \\
\hline
\end{tabular}
\label{tab:datasets_RC26}
\end{center}
\end{table}

\begin{table*}[htbp]
\caption{SPARC Datasets}
\begin{center}
\begin{tabular}{|c|c|c|c|c|}
\hline
\textbf{Dataset}&\textbf{\# of Lanes }&\textbf{Duration (s)}&\textbf{Length (m)}&\textbf{Size (GB)}\\
\hline
3D Point Cloud of LRC 2026 flat & 6 & N/A & N/A  & 0.07 \\ \hline
3D Point Cloud of LRC 2026 inclined & 6 & N/A & N/A  & 0.2 \\ \hline
LRC 2026 \#1 inclined& 6 & 392 & 90 & 12.5 \\ \hline
LRC 2026 \#2 inclined& 3 & 429 & 50 & 13.7 \\ \hline
RoboCup 2026 & 5 & 614 & 100 & 21.2 \\ \hline
\end{tabular}
\label{tab:data_details}
\end{center}
\end{table*}

\section{Open Source Software Stack}
\label{sec:software}
We open-sourced the code base at \url{https://github.com/STAR-Explorer-Robotics/Robocup_system}. The main components of the navigation approach are described below.

\subsection{System Architecture}
\label{sec:software_architecture}
The software architecture of the competition-configured Unitree A2-W is organized as a distributed ROS 2 system comprising onboard computing, robot control, perception, and operator interfaces. The architecture separates computationally intensive perception and interaction functions from the components responsible for real-time robot control and navigation.

\begin{figure}[t]
    \centering
    \includegraphics[width=\linewidth]{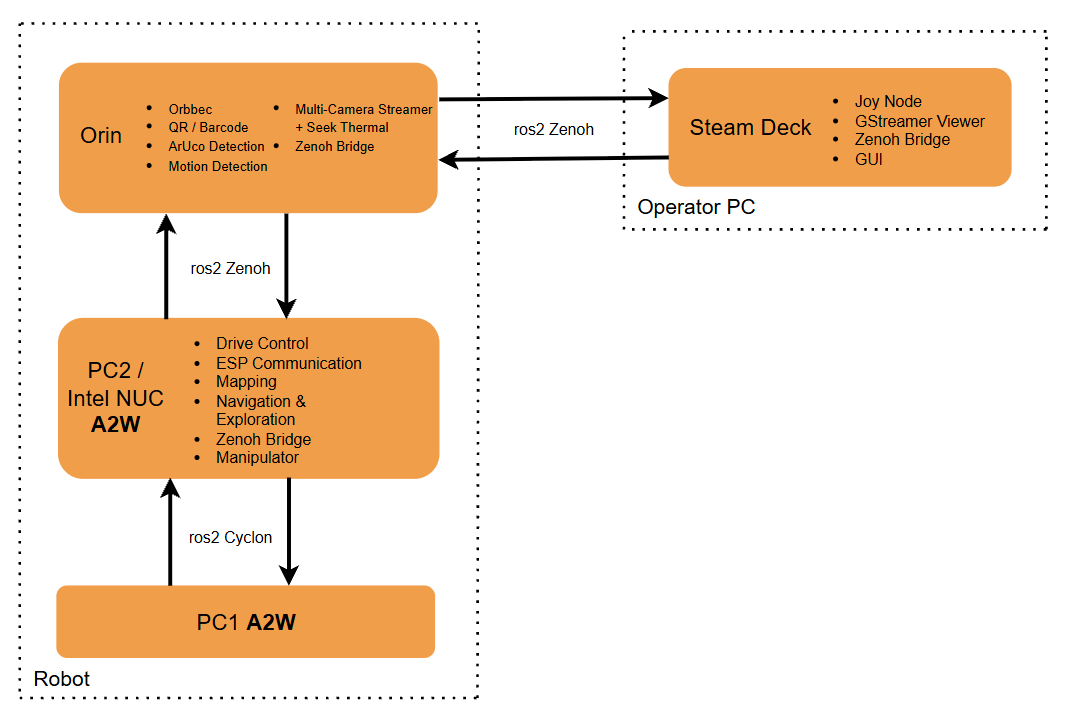}
    \caption{Software and computing architecture of the competition-configured Unitree A2-W. The system is distributed across the robot's onboard computing units and an external operator PC, with ROS 2 and Zenoh providing the communication layer between perception, autonomy, robot control, and operator interfaces.}
    \label{fig:software_architecture}
\end{figure}

As illustrated in Fig. \ref{fig:software_architecture}, the onboard system consists of two computing units, denoted as PC1 and PC2, together with an NVIDIA Jetson Orin platform. PC2, based on an Intel NUC, acts as the main high-level control computer of the A2-W and hosts the core autonomy components, including drive control, mapping, navigation and exploration, ESP-based communication, the Zenoh bridge, and manipulator control. PC1 provides the interface to the underlying A2-W robot system and exchanges robot state and control information with PC2.

The Jetson Orin is primarily used for computationally demanding perception and sensor-processing tasks. These include RGB-D perception, QR and barcode detection, ArUco marker detection, motion detection, multi-camera streaming, and processing of thermal camera data. The perception system provides information to the high-level autonomy components running on PC2.
Communication between the computing components is implemented using ROS 2 with Zenoh as the underlying communication mechanism. The connection between PC1 and PC2 uses ROS 2 CycloneDDS, while the communication between PC2, the Orin, and the operator computer is bridged through Zenoh. This distributed architecture allows sensor processing, robot control, autonomous navigation, and operator interaction to be executed on separate computing resources while maintaining a common ROS 2 interface.
The operator interface is implemented on a dedicated PC running Steam Deck-based interaction software. The operator system provides a graphical user interface, joystick-based control, visualization, and communication with the robot through the Zenoh bridge. In addition, the GStreamer-based video pipeline provides access to multiple camera streams, including thermal imagery, for remote situational awareness.

The resulting architecture separates the system into three functional levels: (i) low-level robot interfacing and control, (ii) onboard perception and autonomous operation, and (iii) operator interaction and situational awareness. This separation supports modular development and allows individual components to be replaced or extended without changing the overall system architecture.

\subsection{The Navigation Strategy}
The autonomous navigation stack is organized around three tightly coupled components: 
(i) robust state estimation and online mapping, 
(ii) terrain-aware path planning, and 
(iii) locomotion execution adapted to the mobility constraints of the Unitree A2-W platform. 
This design reflects the requirements of rescue-robot deployment, where the robot must localize itself in unstructured environments, reason about terrain traversability, and execute feasible motions over ramps, stairs, slopes, gaps, and irregular obstacles.

\paragraph{State estimation and online mapping.}
Robust localization is the foundation of autonomous operation in SPARC environments. 
Unlike structured indoor navigation scenarios, rescue test arenas often contain uneven terrain, repeated geometric patterns, sparse features, and strong motion disturbances caused by legged locomotion. 
To address these challenges, we adopt SuperOdom~\cite{zhao2021super}, an open-source multi-sensor odometry system designed for robust operation in degraded environments. 
SuperOdom uses an IMU-driven tightly coupled estimation framework, in which inertial propagation provides the backbone of the odometry estimate, while LiDAR and camera measurements are incorporated as corrective observations when reliable environmental features are available. 
This design is particularly suitable for legged rescue robots, since the IMU remains continuously available even when visual or geometric features are temporarily degraded.

For scenarios where partial prior knowledge of the test environment is available, we further integrate SuperLoc~\cite{zhao2025superloc}, the follow-up localization framework of SuperOdom. 
SuperLoc allows the robot to exploit a prior map for global localization and drift correction, while still maintaining online odometry for local motion execution. 
Since the onboard LiDARs of the Unitree A2-W are equipped with the latest JT128 modules, we modify the LiDAR pre-processing pipeline to adapt the raw data stream to the input format and timing assumptions of SuperOdom and SuperLoc. 
These modifications include point-cloud format conversion, timestamp handling, and sensor-frame alignment, enabling stable integration with the rest of the navigation stack.

\paragraph{Terrain-aware path planning}
To account for the maneuvering risk of legged locomotion on complex terrain, we extend the TRG-planner~\cite{lee2025trg} as RB-TRG \cite{rb_trg}. 
It constructs a terrain-aware graph from an elevation map. 
Graph nodes are stochastically sampled over the height map, and graph edges are evaluated according to the local terrain geometry between neighboring nodes. 
For each candidate edge, a traversability cost is computed from the expected base attitude of the robot, especially the pitch and roll angles required to traverse the corresponding terrain segment. 
This risk-aware cost allows the planner to prefer routes that are not only short, but also safer for the physical robot to execute.

Based on the generated graph, an $A^*$ search is used to compute a minimum-cost path from the robot's current position to the target location. 
We modify the original TRG formulation and path-finding as RB-TRG \cite{rb_trg}, to better match the mobility characteristics of the Unitree A2-W in the SPARC test environment. 
In particular, we introduce direction-dependent traversal costs to distinguish between uphill and downhill motions, since the stability and achievable velocity of the robot are different when ascending and descending slopes or steps. 
We also tune the edge-validation criteria according to the robot's footprint, body clearance, and empirically observed locomotion limits. 
These adaptations allow the planner to generate paths that are more consistent with the actual capabilities of the platform during competition deployment.

\paragraph{Locomotion Execution}
The planned path is executed through the robot's onboard locomotion interface rather than through direct joint-level control. 
Given a global or local reference path, the navigation module continuously computes velocity commands based on the robot's estimated pose and the local path geometry. 
This closed-loop tracking layer bridges the gap between high-level terrain-aware planning and the proprietary locomotion controller of the Unitree A2-W. 
In practice, this separation allows the planner to focus on selecting safe and feasible routes, while the low-level controller handles gait generation, balance control, and terrain contact adaptation.

\paragraph{Competition Evaluation}
The autonomous navigation stack was evaluated during the LRC in the official competition arena. Autonomous navigation was successfully demonstrated on all evaluated lanes.

During the preliminary round, the terrain modules were placed in the flat configuration. The system achieved a 100\% success rate in the autonomous navigation trials, with an average round-trip completion time of approximately 3~min.

In the later stage of the competition, several terrain modules were inclined to increase the traversal difficulty. Under this configuration, the robot successfully completed 7 out of 10 autonomous navigation trials except the Rotating Ramps, corresponding to a success rate of 70\%. The average round-trip completion time increased to approximately 4--5~min due to the more challenging terrain and reduced traversal speed. No changes were made to the overall navigation architecture between the two configurations. These competition results demonstrate that the integrated localization, terrain-aware planning, and locomotion execution stack can support autonomous navigation in standardized rescue environments, while the reduced success rate on inclined terrain highlights the remaining challenges under more demanding terrain conditions.


\section{Simulation}
\label{sec:simulation}
To facilitate future simulation-based evaluation, we provide a URDF model of the SPARC robot platform and MuJoCo-Lab-compatible terrain assets generated from the collected competition environments~\cite{mjlab}. These assets allow users to load the robot and representative rescue terrains in simulation, replay collected trajectories, and visualize robot states for debugging and analysis.

The simulation integrates an existing open-source locomotion
controller for robot motion execution, while high-level
navigation commands can be generated by external navigation
frameworks. Together with the open-source autonomous
navigation stack presented in Section~V, the simulation
environment allows researchers to develop and test
localization, mapping, path planning, and navigation
strategies in representative rescue scenarios.

\section{Conclusions}
\label{sec:conclusions}

This paper introduced the RoboCup Standard Platform for Autonomous Rescue Competition (SPARC), a new competition of the RoboCup Rescue Robot League based on the Unitree A2-W standard platform. SPARC aims to make research on artificial intelligence for autonomous rescue robots accessible to a broader research community, including groups without expertise in developing custom robot hardware. By combining a capable commercially available robot with established RoboCup Rescue test methods, SPARC enables researchers to focus on autonomous navigation, perception, manipulation, and embodied AI in challenging and reproducible disaster-response scenarios.

To support research and participation in SPARC, this paper presented the first open SPARC datasets collected at the IEEE ICRA 2026 Legged Robot Challenges (LRC) and RoboCup Rescue 2026, an open-source software stack for autonomous navigation, and initial simulation assets implementing the RoboCup Rescue benchmark environments. Together, these resources provide an initial foundation for developing, evaluating, and comparing autonomous rescue systems on a common platform. We hope that SPARC will establish a common benchmark and open-source software ecosystem for autonomous rescue robotics, enabling reproducible research, faster algorithm development, and broader participation from both the robotics and AI communities.



\bibliographystyle{plain}   
\IEEEtriggeratref{7}
\bibliography{references}   

\end{document}